\documentclass[twoside,11pt]{article} 
\usepackage[x11names]{xcolor}
\usepackage{jmlr2e}
\usepackage[frozencache=true, cachedir=minted-cache]{minted} 

\usepackage[french,english]{babel}
\usepackage[utf8]{inputenc}  %
\usepackage[T1]{fontenc}   		%
\usepackage{lmodern} %

\usepackage{caption}
\usepackage{array}
\usepackage{longtable}
\usepackage{arydshln}  %
\usepackage{booktabs} %
\usepackage{graphicx}

\usepackage[top=1in, bottom=1in, left=0.75in, right=0.75in]{geometry}
\usepackage[edges]{forest}
\usepackage{setspace}

\usepackage{bbold}
\usepackage{soul}
\usepackage{yfonts}
\usepackage[shortlabels]{enumitem}
\usepackage{ulem}

\usepackage{float}
\usetikzlibrary{shapes.geometric}
\usetikzlibrary{shapes.arrows}
\usepackage{pgfplots}
\usepgfplotslibrary{groupplots,dateplot}
\usetikzlibrary{patterns,shapes.arrows}
\pgfplotsset{compat=newest}

\usepackage{pifont}%
\newcommand{\cmark}{\ding{51}}

\begin{document}

\title{\texttt{ml\_edm} package: a Python toolkit for Machine Learning based Early Decision Making}

\author{\name Aurélien Renault \email aurelien.renault@orange.com
\AND Youssef Achenchabe \email youssef.achenchabe@... 
\AND Édouard Bertrand \email edouard.bertrand100@gmail.com 
\AND Alexis Bondu \email alexis.bondu@orange.com
\AND Antoine Cornuéjols \email antoine.cornuejols@agroparistech.fr
\AND Vincent Lemaire \email vincent.lemaire@orange.com
\AND Asma Dachraoui \email asma.dachraoui@...
}

\editor{my editor}

\maketitle

\begin{abstract}
    \texttt{ml\_edm} is a Python 3 library, designed for early decision making of any learning tasks involving temporal/sequential data. The package is also modular, providing researchers an easy way to implement their own triggering strategy for classification, regression or any machine learning task. As of now, many Early Classification of Time Series (ECTS) state-of-the-art algorithms, are efficiently implemented in the library leveraging parallel computation. The syntax follows the one introduce in \texttt{scikit-learn}, making estimators and pipelines compatible with \texttt{ml\_edm}. This software is distributed over the BSD-3-Clause license, source code can be found at \url{https://github.com/ML-EDM/ml_edm}. \\
\end{abstract}


\section{Introduction}

In hospital emergency rooms (\cite{mathukia2015modified}), in the control rooms of national or international power grids (\cite{dachraoui2015early}), in government councils assessing critical situations, there is a \textit{time pressure} to make early decisions.
On the one hand, the longer a decision is delayed, the lower the risk of making the wrong decision, as knowledge of the problem increases with time. On the other hand, late decisions are generally more costly, if only because early decisions allow one to be better prepared.
For example, a cyber-attack that is not detected quickly enough gives hackers time to exploit the security flaw found.

A number of applications involve making decisions that optimizes a trade-off between accuracy of the prediction and its earliness. The problem is that favoring one usually works against the other. Greater accuracy comes at the price of waiting for more data. Such a compromise between the \textit{Earliness} and the \textit{Accuracy} of decisions has been particularly studied in the field of Early Classification of Time Series  (ECTS) (\cite{gupta2020approaches}), and introduced by \cite{xing2008mining}.

In this paper, the \texttt{ml\_edm} package is presented, it gathers many state-of-the-art ECTS algorithms in a modular way, clearly differencing the classification part from the triggering part, when possible. Thus, it allows researchers to easily reproduce past results, test well-know methods in different settings and implement new ECTS algorithms.
\section{Implementation details}

\textit{Dependencies: } The \texttt{ml\_edm} package depends on \texttt{numpy} (\cite{van2011numpy}), \texttt{scipy} (\cite{virtanen2020scipy}) and \texttt{pandas} (\cite{mckinney-proc-scipy-2010}) for classic array operations. It also depends on \texttt{scikit-learn} for its API, utilities as well as some classical Machine Learning models. The package is also dependent of \texttt{aeon} (\cite{middlehurst2024aeon}), for some time series specialized features extraction algorithms.

\section{An API for Early Time Series Classification}

One of the main field of research when it comes to Early Decision Making is ECTS. The package has been primarily built to reproduce results from this literature, working only with univariate time series for now (\cite{renault2024early}). In what follows, the different modules are presented through an introductory example. Please note that, as Early Decision Making can be broaden to other ML tasks, the \texttt{ml\_edm} package can also be used to address these. 

\subsection{Cost setting}

One of the keystone of the \texttt{ml\_edm} library is the cost setting, i.e. how much does making a bad prediction costs in comparison of waiting a certain amount of time. Even if these are often difficult to estimate in practice ; still, we argue that those are supposed to act as the \textit{ground truth} used for both training and evaluation. Thus, even when no information about those costs are available we advise to emulate them, using very simple hypotheses, e.g. binary and symmetrical misclassification cost alongside a linear delay cost. In term of API, no \texttt{ml\_edm} estimators can be instantiated/trained without defining some cost setting beforehand using the \texttt{CostMatrices} object.

\begin{minted}
[bgcolor=Snow1, 
baselinestretch=1,
fontsize=\footnotesize,
frame=lines, 
linenos]
{python}
"""
Load time series dataset (X, y) with 
max_T: biggest timestamps
n_classes: number of unique classes in y 
timestamps: set of timestamps that will be of interest (usually a subset of all timesteps)
"""
from ml_edm.cost_matrices import CostMatrices

misclf = 1 - np.eye(n_classes)
def delay(t)
    return t/max_T

cost_matrices = CostMatrices(
    timestamps=timestamps, 
    n_classes=n_classes,
    misclf_cost=misclf,
    delay_cost=delay
)
\end{minted}

\subsection{Classification}

The \texttt{classification} module provides the user an interface to define what will be the strategy to perform classification of varying-length time series in our case. Some simple way of doing this is to use a collection of independently trained classifiers, each of these being specialized for a particular timestep. A \texttt{base\_classifier} to be cloned can be defined, this object can be any of the \texttt{scikit-learn} classifier or pipeline, as long as it has some \texttt{predict\_proba} method. Note that in the case of \textit{end-to-end} models, this classification stage is not mandatory.

\begin{minted}
[bgcolor=Snow1, 
baselinestretch=1,
fontsize=\footnotesize,
frame=lines, 
linenos]{python}
from aeon.datasets import load_classification
from sklearn.ensemble import HistGradientBoostingClassifier
from ml_edm.classification import ClassifiersCollection

X_train, y_train, _ = load_classification("GunPoint", split="train")
X_test, y_test, _ = load_classification("GunPoint", split="test")

# fit as much classifiers as len(timestamps)
collection_clf = ClassifiersCollection(
    base_classifier=HistGradientBoostingClassifier(),
    timestamps=timestamps
)
# a `CostMatrices` object can optionally be passed through the fit method
collection_clf.fit(X_train, y_train) 

for t in timestamps: # eval performances for each estimator within the collection
    accuracy = collection_clf.score(X_test[:, :t], y_test)
    print(f"Accuracy of classifier trained for timestamp no {t}: {accuracy:.4f}")
\end{minted}

\subsection{Trigger module}

Once a cost setting and a classification strategy has been defined, one can safely define some \texttt{EarlyClassifier} estimator to perform ECTS. Here are the different trigger models one can find in the library. Most of them support parallel training, drastically reducing training times.

\begin{table}[!htb]
    \footnotesize
    \centering
    \setlength\extrarowheight{3pt}
    \begin{tabular}{ccc}
        Trigger Model & End2end & Parallel \\
          \hline
          ProbabilityThreshold\footnote{Baseline also implemented in aeon (\cite{middlehurst2024aeon})} & & \cmark \\
          EDSC (\cite{xing2011extracting}) & \cmark & \cmark \\
          ECTS (\cite{xing2012early}) & & \cmark \\
          ECDIRE (\cite{mori2017reliable}) &  & \cmark \\
          Stopping Rule (\cite{mori2017early}) & & \cmark \\
          ECEC (\cite{lv2019effective}) & & \cmark \\
          TEASER (\cite{schafer2020teaser}) & & \cmark\\
          ECONOMY-$\gamma$ (\cite{zafar2021early}) & & \\
          CALIMERA (\cite{bilski2023calimera}) & & \\
          \hline
    \end{tabular}
    \caption{Table of implemented trigger models in \texttt{ml\_edm}.
    }
    \label{tab:refs}
\end{table}

Once fitted, the \texttt{EarlyClassifier} estimator have a custom \texttt{score} function, gradually unveiling the test time series in an online fashion. By default, three classical evaluation metrics are outputed, i.e. the \textit{Average cost}, the \textit{Accuracy} as well as the \textit{Earliness}.

\begin{minted}
[bgcolor=Snow1, 
baselinestretch=1,
fontsize=\footnotesize,
frame=lines, 
linenos]{python}
from ml_edm.early_classifier import EarlyClassifier
from ml_edm.trigger import EconomyGamma

early_clf = EarlyClassifier(
    chronological_classifiers=collection_clf, 
    trigger_model=EconomyGamma(), 
    cost_matrices=cost_matrices,
    #prefit_classifiers=True when a classification module has already been fitted
)
early_clf.fit(X_train, y_train)

avg_cost, accuracy, earliness = early_clf.score(X_test, y_test)
print(f" Average cost: {avg_cost} \n Accuracy: {accuracy} \n Earliness: {earliness}")
\end{minted}

\section{Conclusion}

\texttt{ml\_edm} is a Python package implementing the main state-of-the-art ECTS algorithms. An emphasize is made on the \textit{cost-sensitive} aspect of the task. Future works include extension to other \textit{Early Decision Making} tasks beside classification, as well as handling multivariate and irregular time series.

\bibliography{ref}

\end{document}